\documentclass[a4paper,twoside]{article}
\usepackage{epsfig}
\usepackage{subcaption}
\usepackage{calc}
\usepackage{amssymb}
\usepackage{amstext}
\usepackage{amsmath}
\usepackage{amsthm}

\usepackage{pslatex}
\usepackage{apalike}
\usepackage{SCITEPRESS}     

\begin{document}
\title{Effect of Word-Embedding Models on Hate and Offensive Speech Detection}
\author{\authorname{Safa Alsafari \sup{1}, Samira Sadaoui\sup{1} and Malek Mouhoub \sup{1}}
\affiliation{\sup{1}University of Regina}

}


\abstract{Deep learning was adopted successfully in hate speech detection problems, but very minimal for the Arabic language. Also, the word-embedding models\' effect on the neural network's performance were not adequately examined in the literature. Through 2-class, 3-class, and 6-class classification tasks, we investigate the impact of both word-embedding models and neural network architectures on the predictive accuracy. We first train several word-embedding models on a large-scale Arabic text corpus. Next, based on a reliable dataset of Arabic hate and offensive speech, we train several neural networks for each detection task using the pre-trained word embeddings. This task yields a large number of learned models, which allows conducting an exhaustive comparison. The experiments demonstrate the superiority of the skip-gram models and CNN networks across the three detection tasks.}

\onecolumn \maketitle \normalsize \setcounter{footnote}{0} \vfill

\section{\uppercase{Introduction}}
\label{sec:introduction}
\noindent Hate and offensive speech detection have become a popular research area in recent years. There are two main approaches to detect hateful/offensive content.  The first approach extracts ngram features from texts explicitly, either word ngram \cite{Davidson2017,Burnap2015} or characters ngram \cite{Mehdad2016,Waseem2016,Malmasi2018}, and then applies suitable learning algorithms to the features. This method can produce relatively high performance. However, it captures only short-range dependencies between words and cannot model long-distance dependencies between words with the context size being limited to a fixed number of tokens. Moreover, this method suffers from data sparseness, especially for languages with rich morphology, like Arabic.
In contrast,  the second approach adopts an end-to-end classification pipeline that implicitly extracts features from raw textual data using neural network algorithms. The latter consists of at least three layers: input, hidden, and output. The input layer transforms the words via an embedding layer into one-dimensional vectors before passing them to the hidden layer. These vectors can be initialized randomly or using learned word embedding.

Previous studies examined the effects of word-embedding models on the performance of text classification systems \cite{Ghannay2016,Wang2018,STEIN2019216,Elrazzaz2017}. Nevertheless, no work compared the impact of different word embeddings on hate speech detection. We focus on Arabic, as studies are minimal for this language.  Elrazzaz et al. \cite{Elrazzaz2017} analyzed the effect of four pre-trained word embeddings using intrinsic and extrinsic evaluations. The authors showed that word embedding'\ s performance can be significantly affected by the size and quality of the corpus used to learn the embedding. In our study, to mitigate this effect and to fairly compare the word-embedding frameworks, we train all them on the same background corpus and assess their performance on challenging hate speech detection tasks. Across languages, hate speech detection was commonly modeled as a binary classification problem where the goal is to discriminate between clean text and text with hateful utterance. Sometimes the problem is extended to three classes (clean, hate, and offensive), but not to more fine-grained categories. 

\medskip

Our study contributes to the literature in several ways. First, we train five word-embeddings using a very vast Arabic corpus. Rather than using only Arabic Wikipedia to train the embeddings, which is the standard practice in existing pre-trained word-embedding models, we employ 10 Gigabyte corpus that we created by combining Wikipedia, United Nations monolingual, and a Twitter corpus.  To the best of our knowledge, this is the largest dataset used for training Arabic word embeddings. We also experiment with the random initialization of word embedding and use it as a baseline in our analysis.  We made all the pre-trained embedding models publicly available on Github. Second, we conduct an intensive empirical investigation of the effect of the pre-trained word embeddings on the deep learning performance using a reliable Arabic hate speech dataset developed recently in \cite{Alsafari2020}. For this purpose, we adopt four neural network architectures, Convolutional Neural Network (CNN), Bidirectional Long Short-Term Memory (BILSTM), Gated Recurrent Unit (GRU), and the hybrid CNN+BILSTM.  Third, we train and then evaluate the performance of 24 pairs of word embedding and deep neural network for each of the three classification tasks:
\begin{itemize}
    \item 2-class (clean vs. hate/offensive)
\item 3-class (clean vs. hate vs. offensive)
\item 6-class (clean vs. offensive vs. religious hate vs. nationality hate vs. ethnicity hate vs. gender hate)
\end{itemize}
As a result, we develop a tally of 144 hate speech classifiers,  which allows us to conduct an exhaustive comparison.  To account for the stochastic nature of neural networks, we train and test each of the 114 classifiers 15 times, each with the same hyper-parameter values but with different random weight initialization. We then average their predictive accuracy.

\medskip

The rest of this paper is structured as follows.  Section 2 describes recent studies on deep learning for hate/offensive speech detection. Section 3 presents five word-embedding models, and Section 4, the four deep neural network approaches. Section 5 exposes the experimental framework, including the pre-training of the word embeddings, our Arabic hate/offensive speech dataset, and the training of the neural networks. Section 6 reports and discusses the performance results for the three detection tasks. Finally, Section 7 summarizes our research findings.

\section{\uppercase{Related Work}}
\noindent Recently, there has been much interest in adopting neural networks to hate speech and offensive language detection on social media.  Traditional classification methods rely on feature engineering and manually convert texts into feature vectors before classifying them with standard algorithms, such as Support Vector Machines (SVM) and Naive Bayes.  In contrast, neural networks can automatically learn the representations of input texts with different levels of abstraction and subsequently use the acquired knowledge to perform the classification task. The two most popular neural network architectures adopted in the field of hate speech detection are the Convolutional Neural Networks (CNNs) \cite{Gamback2017,Park2017,Zhang2018a,Zampieri2019} and Recurrent Neural Networks (RNN), such as Gated Recurrent Unit (GRU) \cite{Zhang2018a,Zhang2018b} and  Long Short Term Memory (LSTM) \cite{Badjatiya2017,DelVigna2017,Pitsilis2018}.

Based on an English Twitter hate speech dataset developed in \cite{Waseem2016}, the authors in \cite{Gamback2017} trained a CNN architecture using random and word2vec word-embedding models to detect racism and sexism tweets. Additionally, they experimented with CNN trained on character n-grams and then a combination of word2vec and  character n-grams. The performance of the word2vec+CNN model outperformed the other models in terms of the F-macro metric.

The authors in \cite{Park2017} experimented with three models to detect abusive language: word-based CNN, character-based CNN, and hybrid CNN that takes both words and characters as inputs.
The binary classification results showed that the hybrid approach outperforms the word and char-based models as well as traditional classifiers, such as Logistic Regression (LR) and SVM.

In \cite{Badjatiya2017}, the authors leveraged LSTM with random word embedding to learn feature vectors, which is then utilized by a Gradient Boosted Decision Trees (GBDT) classifier for the task of hate speech detection. The experiments demonstrated that this ensemble yields the best accuracy compared to the ensemble of SVMs or GBDTs with traditional word n-gram as well as the ensemble of LRs with character n-gram. The proposed method also outperforms single deep learning classifier, such as CNN and LSTM.

Another work \cite{Pitsilis2018} devised an ensemble of three and five LSTM classifiers to solve the task of detecting racism, sexism, and neutral Tweets. Each classifier is trained using vectors of user-related information and word frequency. Lastly, the results of the classifiers are aggregated using voting and Confidence approaches.The ensemble classification results produced an overall improvement over previous state-of-the-art approaches, especially for short text classification.

Zhang et al.\cite{Zhang2018a} trained a combined CNN and GRU model on several public hate and abusive language detection datasets. The results of this ensemble approach achieved an improvement of 2\% to 9\%  in the detection accuracy compared to the state-of-the art.
In the follow-up work \cite{Zhang2018b}, an
extension of CNN model (skipped CNN) was suggested. The extension captures the implicit discriminative features better. It adopts a skipped CNN layer in addition to the regular convolution layer to extract skip-gram like features. The experimental analysis indicated that the new model is more successful in detecting hateful content.

\section{\uppercase{Word-Embedding Models}}
\noindent Word embedding is a way of mapping words into fixed, dimensional, real vectors that capture both the semantic and syntactic information regarding the words. These word embeddings are used for initializing weights of the first layer of a neural network, and therefore its quality has a significant effect on the predictive network performance.  Word embeddings are often trained on massive unlabelled corpora, such as Wikipedia.  Even though there are several pre-trained word embeddings for the Arabic language \cite{grave2018learning,Mohammad2017},  they were trained on different background datasets, which makes it very difficult to compare the effects of their training frameworks from a logical viewpoint.  Consequently, in our work, to fairly assess the effectiveness of different word-embedding models, we build a large-scale Arabic corpus (described in Section 5.1) to train five embedding learning frameworks described below.  

 \paragraph{\bf Glove:}
 This model learns word embedding based on the corpus statistics and co-occurrence word counts matrix \cite{Pennington2014}. Word embedding is extracted based on the ratios of word-word co-occurrence probabilities that are believed to encode some meaning. 
 The objective function of the glove model is as follows\cite{Pennington2014}:
\begin{equation} \label{eq1}
\frac{1}{2}\sum_{i,j=1}^{V}f(X_{ij})(w_{i}^{T}\tilde w_{j}+b_{i}+\tilde b_{j}-log X_{ij})^{2}
\end{equation}
where V is the vocabulary, $f(X_{ij})$ the weighting function to discount the effects of rare and large co-occurrence frequencies, $b_{i}$ and $b_{j}$ bias terms, and $X_{ij}$ the word-context matrix count.

\paragraph{\bf Word2vec: } This model is one of the early neural network-based frameworks for creating word embedding \cite{Mikolov2013k}. In our study, we adopt two variants of Word2vec:  Continuous Bag-Of-Words (CBOW) and Sikp-gram.  The CBOW model learns word embedding by training a feed-forward neural network using word-context pair information with language modeling objective, where the goal is to predict the word giving its context. 
 The objective of the CBOW model is to maximize the following average log likelihood probability: 
\begin{equation} \label{eq2}
\frac{1}{t}\sum_{t=1}^{T}\log  p(w_{t}|w_{c}) 
\end{equation}
 where w is the target word, and $w_{c}$ represents the sequence of words in context. 
 
On the other hand, the Skip-gram model aims to predict the context word giving the word with the following objective function \cite{Mikolov2013k}: 
 \begin{equation} \label{eq3}
\frac{1}{t}\sum_{t=1}^{T}\sum_{c \in C_{t}}\log  p(w_{c}|w_{t}) 
\end{equation}
where $c_{t}$ denotes the set of context indices of word $w_{t}$. 
 Both models use negative sampling and hierarchical softmax algorithms for a more efficient handling of frequent and rare words.

\paragraph{\bf FastText: } This model \cite{grave2018learning}
 is an improvement of the Word2vec in which the embeddings are created by incorporating character n-grams  information. It first learns an embedding for each character n-grams, and then 
 the embedding of an individual word  is computed as the sum of its character ngrams embedding.
 This model is especially effective for morphologically rich language,  such as Arabic, because it learns sub-word information and makes it possible to create representations for rare or out of vocabulary words.
 Similar to Word2vec, FastText embedding can be trained using CBOW or skip-gram model. We experiment with both variants in our work.

\section{\uppercase{Neural Network Architectures}}
\noindent For our hate and offensive speech detection application, we compare four neural network architectures, CNN, BILSTM, GRU and the hybrid CNN+BILSTM. In this section, we describe their underlying architectures and configurations. 
 
 \paragraph{\bf CNN: }
Our basic convolutional neural network is similar to the one developed in \cite {Kim2014}.
This network takes the output features of the word embedding models as inputs and applies one-dimensional convolution layer to the features to learn context information of the words. This layer consists of 250 filters of size 2. It is followed by a dropout with a rate of 0.5, and then by the max pooling function.
After that, a linear fully connected layer is used to output the probability distribution over the target classes; the class with the highest probability is selected as the final label.
  
\paragraph{\bf LSTM: }
This network is a refinement of the general RNN architecture with the ability to capture long-term dependency information. LSTMs deal with the inputs sequentially, which makes it suitable for textual data. It processes one word at a time by passing it through LSTM units. Each unit takes as inputs the embedding vector of the current word and the output from preceding unit, and uses these inputs to update its internal memory cell, thus recursively accumulates information about all other words in the text.  For our hate speech detection problem, we adopt a bidirectional LSTM (BILSTM) network by using two LSTMs that process the texts from left to right, and vice versa.  The output of both networks is regularized by a dropout layer with a rate of 0.5. The regularized outputs are then concatenated and mapped to a final label by using a fully connected layer that produces the probability distribution over the labels.
 
\paragraph{\bf GRU: }
This network is another variation of the RNN, proposed by Cho et al. \cite{Cho2014}. It is faster to train than LSTM and with lesser computational load. A GRU unit has a simple architecture compared to the LSTM, and according to Jozefowicz et al \cite{Jozefowicz2015}, GRU is able to outperform LSTM on several classification tasks. 
We experiment with a simple GRU model that possesses 100 hidden units in the GRU layer and a 0.5 dropout layer that is followed by a final fully connected layer.

\paragraph{\bf CNN+BILSTM: }
This hybrid model is composed of CNN and BILSTM networks. It consists of CNN front-end with 250 hidden units in the convolutional layer using a kernel size of 2 and a dropout rate of 0.5. The subsequent layer is the max pooling followed by a BILSTM layer with 250 hidden units. The BLSTM outputs are then fed to a fully connected layer that returns the label with the highest probability.

\medskip 

For all these network architectures, we utilize the activation function ReLu for the convolution layer, and Adam solver for the parameter optimization with an initial learning rate of 0.0001 (this rate decays by a factor of 0.5) and a batch size of 32. 
For the binary classification task, we adopt the Sigmoid activation function for the fully connected layer along with the binary cross entropy loss function. For the multi-class classification settings, we use the Softmax function and the categorical cross entropy loss. 

\section{\uppercase{Experiment Setup}}
\subsection{Word-Embedding Corpus}
We first develop five word-embedding models: Glove, Word2vec CBOW (w2v-cb), Word2vec Skip-gram (w2v-sg), Fasttext CBOW (ft-cb) and Fasttext Skip-gram (ft-sg). More precisely, we train these models on a collection of Wikipedia dump of 3 million Arabic sentences, an united nation corpus of 9.9 million Arabic sentences, and a tweet corpus of 6.5 million Arabic tweets.  In fact, we collect the tweets specifically to build a corpus that is closely related to the domain of hate speech. The entire combined corpus consists of 19.4 million sentences with 0.5 billion tokens. Hence, we train the five word embeddings with this very large unlabeled Arabic textual corpus.  For training, we use the parameter values proved efficient in the literature: 300 for the dimension size of word vector and 5 for the window context size. We set the minimum word count to 5, which results into a vocabulary size of 1.1 million words.

\begin{table}[h]
\caption{2-Class Distribution.}
\label{tab2C:CorpusDistribution}
\centering
\resizebox{0.5\columnwidth}{!}{%
\begin{tabular}[t]{|l|c|}
\hline 
\multicolumn{1}{|c|}{Label} & Tweets \\ \hline 
Clean                       & 3480   \\ \hline
Offensive/Hate              & 1860   \\ \hline 
Total                       & 5340   \\ \hline 
\end{tabular}
}
\end{table}

\begin{table}[h]
\caption{3-Class Distribution.}
\label{tab3C:CorpusDistribution}
\centering
\resizebox{0.5\columnwidth}{!}{%
\begin{tabular}[t]{|l|c|}
\hline 
\multicolumn{1}{|c|}{Label} & Tweets \\ \hline
Clean                       & 3480   \\ \hline
Offensive                   & 437    \\ \hline 
Hate                        & 1423   \\ \hline 
Total                       & 5340   \\ \hline
\end{tabular}
}
\end{table}
\begin{table}[h]
\caption{6-Class Distribution.}
  \label{tab6C:CorpusDistribution}
  \centering
\resizebox{0.5\columnwidth}{!}{%
\begin{tabular}[t]{|l|c|}
\hline
\multicolumn{1}{|c|}{Label} & Tweets \\ \hline
Clean                       & 3480   \\ \hline
Offensive                   & 437    \\ \hline
Religious  Hate             & 321    \\ \hline
Gender Hate                 & 352    \\ \hline
Nationality Hate            & 368    \\ \hline
Ethnicity Hate              & 382    \\ \hline
Total                       & 5340   \\ \hline
\end{tabular}
}
\end{table}

\subsection{Hate and Offensive Speech Dataset}
We employ a reliable Arabic hate speech multi-class dataset developed recently in the work~ \cite{Alsafari2020} to perform our empirical comparison. The authors retrieved textual data from Arabic Twitter over six months using several searching strategies, such as content words (e.g., Women and Arab), user-generated hashtags (e.g., \# Muslims and \# Kurds) {and public figures timeline}. After a rigorous cleansing, the final dataset possesses 5360 tweets annotated by three Arabic native speakers using three hierarchical annotation levels \cite{Alsafari2020}, summarized below:

\bigskip

\noindent {\bf Two-class Labeling: } At the first level (Table \ref{tab2C:CorpusDistribution}), the posts were labeled as either Clean or Offensive/Hate in case they use hurtful or socially unacceptable language:
        \begin{itemize} 
        \setlength{\itemsep}{0pt}
            \item {Clean:} posts not containing offensive, hateful language and profanity.
            \item {Offensive/Hateful:} posts containing non-acceptable, offensive or hateful language, such as insults, threats, profanity or swear words. 
        \end{itemize}
        
\noindent {\bf Three-class Labeling:} At the second level (Table \ref{tab3C:CorpusDistribution}), Offensive/Hate posts were further categorized as either Hate or Offensive according to whether they attack people based on the protected characteristics, such as religion and nationality: 
        \begin{itemize}
            \item {Offensive-NotHateful:} posts containing non-acceptable language or general profanity but are not attacking people based on their protected characteristics.
            \item {Hateful:} posts containing insults/threats/irony to an individual or a group based on the protected characteristics. 
        \end{itemize}
 
\noindent {\bf Six-class Labeling:}  The third level (Table \ref{tab6C:CorpusDistribution}) is a more fine grained classification of hateful tweets where the goal is to further classify the hateful tweets as either religious hate, gender hate, nationality hate or ethnicity hate: 
        \begin{itemize}
            \item {Religion-based: } This type of hate attacks people based on their beliefs or religious background. The speech targets people belonging to different sects and branches of any religion. 
            \item {Gender-based: } This hate type also known as sexism is increasingly common on social networks. This hate speech is motivated by bias against a person’s gender with women being disproportionately targeted. It is becoming prominent on Arabic social media as a result of several recent movements toward gender equality and women empowerment.
            \item {Nationality-based: } This type of hate speech is motivated by bias against a person’s nationality, which is becoming increasingly common on Arabic social media as a result of recent conflicts in the middle east regions.
            \item {Ethnicity-based: } This type involves hate speech along ethnic lines and derogatory remarks about other tribes, races and communities. The study of this type of hate is significantly important in the Arabic region due to various regional and tribal groupings.
        \end{itemize}
        
\smallskip

At each level, the labeling quality was assessed using the Fleiss' Kappa metric \cite{Alsafari2020}. The results for all the levels fall into the interval [0.81, 0.99],  which corresponds to  "{\it almost perfect agreement}” \cite{kappastatistic}. Moreover, we pre-processed the Arabic dataset by removing punctuation, diacritic markers and website URLs. We replaced all the digits with the number “99”, users'\ mentions with "UserMention", and all emojis and symbols with "Emojis".  We additionally normalized hashtags by discarding the \# symbol and separating it into its constituent words, and elongated words by deleting letter repetition of more than two.

Table \ref{tab6C:CorpusDistribution} presents the final Arabic hate speech corpus. As observed, the dataset is typically imbalanced with respect to the hate and offensive classes.

\begin{table*}[t]
\centering
\footnotesize
\caption{Two-Class Performance Results.}
\begin {tabular}{|l|c|c|c|c|c|c|c|c|c|c|c|c|}
\hline
 & \multicolumn{3}{c|}{CNN}       & \multicolumn{3}{c|}{GRU}       & \multicolumn{3}{c|}{BILSTM}    & \multicolumn{3}{c|}{Hybrid} \\ \cline{2-13} 
                  & P     & R     & \multicolumn{1}{|c|}{\begin{tabular}[c]{@{}c@{}}FM\end{tabular}}          & P     & R     & \multicolumn{1}{|c|}{\begin{tabular}[c]{@{}c@{}}FM\\\end{tabular}}          & P     & R     & \multicolumn{1}{|c|}{\begin{tabular}[c]{@{}c@{}}FM\end{tabular}}          & P      & R     & \multicolumn{1}{|c|}{\begin{tabular}[c]{@{}c@{}}FM\end{tabular}}          \\ \hline 
Rand            & 82.36 & 85.44 & \textbf{83.53}          & 79.62 & 82.15 & 80.48          & 80.45 & 84.23 & 81.78          & 80.50  & 82.85 & 81.30          \\ \hline 
w2vcb     & 84.56 & 87.40 & 85.56          & 83.50 & 88.39 & 85.16          & 84.92 & 88.50 & \textbf{86.21} & 84.69  & 87.55 & 85.68          \\ \hline 
w2vsg       & 86.23 & 88.84 & \textbf{87.22} & 82.46 & 89.80 & 84.62          & 82.16 & 89.19 & 84.14          & 85.33  & 88.47 & 86.42  \\ \hline
ftcb           & 78.92 & 81.35 & 79.47          & 81.79 & 86.11 & 83.10          & 82.01 & 85.67 & \textbf{83.26}          & 80.30  & 85.77 & 81.80          \\ \hline  
ftsg             & 86.09 & 87.97 & \textbf{86.78}          & 83.76 & 88.48 & 85.30 & 82.25 & 89.30 & 84.34          & 85.12  & 87.64 & 85.99          \\ \hline 
Glove             & 84.28 & 87.24 & \textbf{85.39}          & 80.77 & 87.88 & 82.79          & 80.68 & 88.32 & 82.80          & 82.93  & 85.99 & 84.02          \\ \hline
\end {tabular}  \label{tab1:Two-Class-Classification}
\end{table*}

\begin{figure*}[t]
\begin{center}
  \includegraphics[width=120mm,height=40mm,scale=0.5]{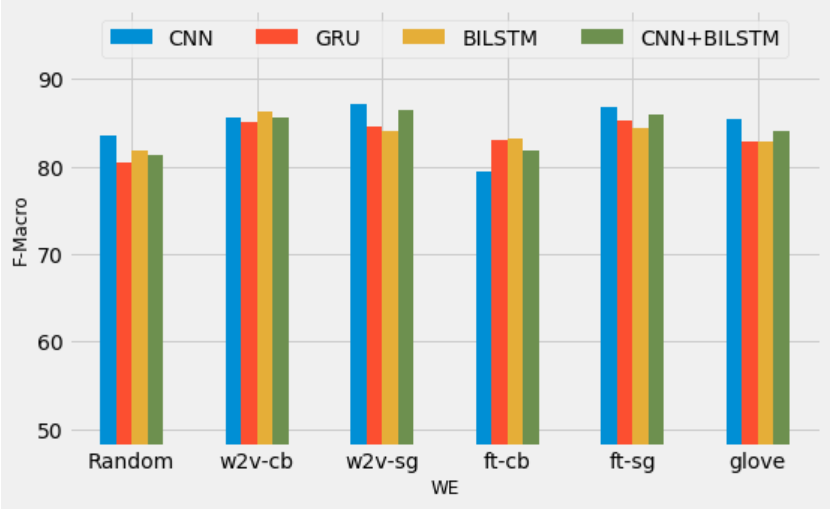}
  \caption{Word Embedding Performance for the 2-class Dataset.}
   \label{fig2:Two-Class-Classification}
\end{center}
\end{figure*}

\subsection{Neural Network Training}
We consider three different prediction tasks for hate and offensive speech detection: 
\begin{itemize}
\item {\it Two classes:} This is a binary classification problem where every tweet is classified as either Clean or Offensive/Hate. 
\item {\it Three classes:} In this classification task, hate tweets are distinguished from  offensive tweets. 
\item {\it Six classes:} Here, hateful tweets are further sub-categorized according to the targeted group, including gender, religious, nationality or ethnicity. 
\end{itemize}

\begin{table*}[]
\caption{Three-Class Performance Results.}
\centering
\footnotesize
\begin{tabular}{|l|c|c|c|c|c|c|c|c|c|c|c|c|}
\hline
 & \multicolumn{3}{c|}{CNN}       & \multicolumn{3}{c|}{GRU}       & \multicolumn{3}{c|}{BILSTM}    & \multicolumn{3}{c|}{Hybrid} \\ \cline{2-13} 
                  & P     & R     & \multicolumn{1}{|c|}{\begin{tabular}[c]{@{}c@{}}FM\end{tabular}}       & P     & R     & \multicolumn{1}{|c|}{\begin{tabular}[c]{@{}c@{}}FM\end{tabular}}          & P     & R     & \multicolumn{1}{|c|}{\begin{tabular}[c]{@{}c@{}}FM\end{tabular}}          & P      & R     & \multicolumn{1}{|c|}{\begin{tabular}[c]{@{}c@{}}FM\end{tabular}}          \\ \hline 
Rand   & 67.89 & 75.04 & \textbf{70.73} & 61.64 & 69.92 & 64.31 & 64.91 & 75.04 & 68.27          & 64.88 & 72.94 & 67.68 \\ \hline
w2vcb & 71.30 & 79.08 & \textbf{74.12} & 69.61 & 78.58 & 72.82 & 71.16 & 78.65 & 74.07          & 69.48 & 77.16 & 71.60 \\ \hline 
w2vsg   & 72.22 & 79.82 & \textbf{75.16} & 70.70 & 80.53 & 73.99 & 68.94 & 79.70 & 72.38          & 71.41 & 79.53 & 73.83 \\ \hline 
ftcb  & 64.23 & 70.70 & 66.06          & 66.52 & 76.15 & 69.77 & 67.74 & 75.18 & \textbf{70.46} & 61.44 & 74.21 & 63.95 \\ \hline 
ftsg    & 73.69 & 80.37 & \textbf{76.09} & 71.86 & 80.54 & 75.00 & 72.03 & 79.87 & 74.56          & 72.63 & 78.40 & 74.37 \\ \hline
Glove    & 69.89 & 79.65 & \textbf{73.36} & 68.99 & 79.95 & 72.74 & 69.28 & 79.62 & 72.74          & 69.08 & 76.80 & 71.67 \\ \hline
\end{tabular}
\label{tab3:Three-Class-Classification}
\end{table*}

\begin{figure*}[]
\begin{center}
  \includegraphics[width=120mm,height=40mm,scale=0.5]{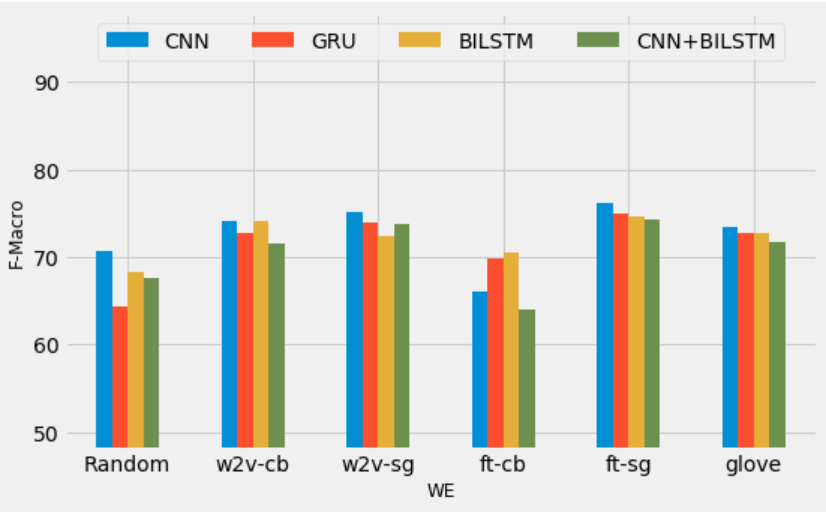}
  \caption{Word Embedding Performance for the 3-class Dataset.}
   \label{fig3:Three-Class-Classification}
\end{center}
\end{figure*}

\begin{table*}[]
\caption{Six-Class Performance Results.}
\centering
\footnotesize
\begin{tabular}{|l|c|c|c|c|c|c|c|c|c|c|c|c|}
\hline 
& \multicolumn{3}{c|}{CNN}       & \multicolumn{3}{c|}{GRU}       & \multicolumn{3}{c|}{BILSTM}    & \multicolumn{3}{c|}{Hybrid} \\ \cline{2-13} 
                  & P     & R     & \multicolumn{1}{|c|}{\begin{tabular}[c]{@{}c@{}}FM\end{tabular}}          & P     & R     & \multicolumn{1}{|c|}{\begin{tabular}[c]{@{}c@{}}FM\end{tabular}}          & P     & R     & \multicolumn{1}{|c|}{\begin{tabular}[c]{@{}c@{}}FM\end{tabular}}         & P      & R     & \multicolumn{1}{|c|}{\begin{tabular}[c]{@{}c@{}}FM\end{tabular}}         \\ \hline
Rand  & 62.66 & 72.42 & \textbf{66.32} & 36.64 & 42.66 & 36.81          & 53.98 & 66.02 & 57.99 & 55.96 & 63.92 & 58.28          \\ \hline 
w2vcb & 63.74 & 74.00 & 67.63          & 60.40 & 75.12 & 65.99          & 61.71 & 74.71 & 66.80 & 64.23 & 73.80 & \textbf{67.75} \\ \hline 
w2vsg   & 66.77 & 76.84 & \textbf{70.80} & 59.91 & 75.06 & 65.36          & 55.52 & 73.19 & 61.40 & 66.93 & 74.82 & 69.83          \\ \hline 
ftcb  & 51.02 & 62.66 & 53.53          & 49.76 & 69.22 & \textbf{55.59} & 48.81 & 66.99 & 54.53 & 45.20 & 60.74 & 49.12          \\ \hline 
ftsg    & 68.32 & 76.47 & \textbf{71.06} & 63.12 & 78.19 & 68.84          & 58.32 & 75.10 & 63.92 & 65.35 & 77.37 & 69.71          \\ \hline 
Glove    & 64.66 & 76.19 & \textbf{69.04} & 58.51 & 78.06 & 65.37          & 57.95 & 77.36 & 64.76 & 62.40 & 72.95 & 66.16          \\ \hline 
\end{tabular}
\label{tab1:Six-Class-Classification}
\end{table*}

\begin{figure*}[]
\begin{center}
  \includegraphics[width=120mm,height=40mm,scale=0.5]{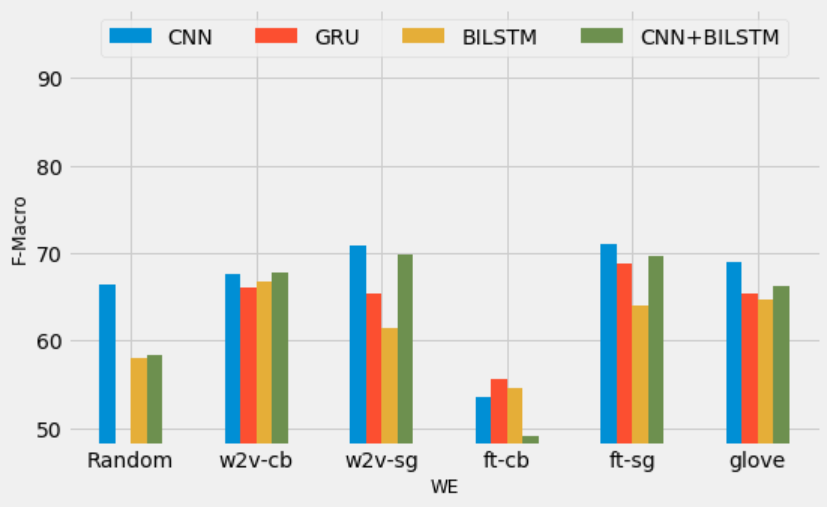}
  \caption{Word Embedding Performance for the 6-class Dataset.}
   \label{fig3:Six-Class-Classification}
\end{center}
\end{figure*}

We partition our corpus into training, validation, and testing subsets containing 60\%, 10\%, and 30\% of instances. We also initialize the four deep neural networks with random word-embedding and use its classification results as a baseline for the analysis and comparison. For each detection task, we train 24 pairs of word embedding/deep neural network (6 word-embeddings x 4 neural networks). Consequently, we produce a tally of 114 hate speech detection models: 24 binary classifiers, 24 3-class classifiers, and 24 6-class classifiers.  

\smallskip

To assess whether the experimental results were systematic or caused by the stochastic nature of neural network training, we train each of the 114 classifiers 15 times, each time with the same set of hyper-parameter values but with different random weight initialization. Each classifier's final accuracy is obtained using the holdout dataset and macro averaging the Precision (P), Recall (R), and F-macro (FM) values.  These metrics are suitable for imbalanced data scenarios.

\section{\uppercase{Results and Discussion}}
\noindent For the binary prediction task, Table \ref{tab1:Two-Class-Classification} and Fig \ref{fig2:Two-Class-Classification} expose the accuracy results of the classifiers trained with random and the five pre-trained word embedding models. The best result achieved by each embedding is highlighted in bold. The results unveil a number of interesting effects the word embeddings have on hate speech detection. First, it can be seen that using pre-trained word embedding has a positive effect on the performance compared to random word-embedding.
Performance is the highest when using CNN with word2vec skip-gram with 0.44\% F-Macro improvement over the second best model (CNN with fasttext skip-gram), and 3.7\% over the baseline classifier.
This model also provides the best Precision and Recall values (which is also true for the two other detection tasks). For the neural networks, we can see the superiority of the CNN, especially when combined with skip-gram based word-embedding over the RNN-based model and CNN+BILSTM-based model.

\smallskip

The performance results for the three-class classification task are presented in Table \ref{tab3:Three-Class-Classification} and Fig \ref{fig3:Three-Class-Classification}.
Similar to our previous findings, overall, the best classifier-embedding pair is CNN with fasttext skip-gram,  which achieved 0.93\% F-Macro improvement over the second best model (CNN with fasttext skip-gram, and 5.4\% over the baseline model. For the six-class classification task, it can be seen from Table \ref{tab1:Six-Class-Classification} and Fig \ref{fig3:Six-Class-Classification} that the optimal classifier-embedding pair is again the CNN using the fasttext skip-gram embedding with 0.26\%  F-Macro improvement over the second best model (CNN with fasttext skip-gram), and 4.7\% over the baseline model.

\smallskip

\noindent{\bf A Summary:} Despite being trained on the same corpus, we found differences between word-embedding performances consistently across the three detection tasks.  We conclude that the CNN architecture trained with skip-gram word-embedding seems to be particularly well suited for the purpose of hate speech detection. Another interesting finding was that the
Fasttext skip-gram embedding always outperforms the Fastext CBOW model when using the same neural network architecture across all the classification tasks. Furthermore, unlike the other embeddings, Fasttext-CBOW works best when combined with RNN (both GRU and BILSTM) networks. The results demonstrated that while the network architecture is essential, the word-embedding models are equally crucial for hate speech detection.

\section{\uppercase{Conclusion and Future Work}}
\noindent Our study investigated how the word-embedding framework affects deep neural networks in the context of hate and offensive speech detection. We trained and compared five word-embedding models using various neural network architectures, including CNN, GRU, BILSTM, and hybrid CNN+BILSTM. Based on an Arabic hate speech dataset, we assessed the performance of each word embedding-classifier pair for three classification tasks, 2-class, 3-class, and 6-class. The results showed that skip-gram models produced more eﬀective representations than other word embeddings. In terms of neural networks, CNN is the best performing across the three classification tasks. 

\medskip

We would like to considerably increase the Arabic hate speech corpus' size to improve even better predictive accuracy for future work.  We are currently scraping millions of textual data from Twitter. However, annotating this huge textual corpus is very challenging. To address the data labeling problem, we will explore two beneficial methods:  1) semi-supervised classification approaches that learn from a few labeled data
(our initial annotated dataset) along with many unlabeled data \cite{elshaar2020semi},  and 2) incremental classification approaches that adjust a classifier progressively with new data \cite{AnowarFlairs2020},\cite{SMC2020}.

\bibliographystyle{apalike}
{\small
\bibliography{example}}
\end{document}